\documentclass[10pt,twocolumn,letterpaper]{article}

\usepackage{cvpr}
\usepackage{times}
\usepackage{epsfig}
\usepackage{graphicx}
\usepackage{amsmath}
\usepackage{amssymb}
\usepackage{tabularx}
\usepackage{multirow}
\usepackage{hhline}
\usepackage{subfig}
\usepackage{booktabs}
\usepackage{sidecap}
\usepackage{rotating}
\usepackage{lipsum}
\usepackage[usenames, dvipsnames]{color}

\usepackage[breaklinks=true,bookmarks=false]{hyperref}

\def\ie{\emph{i.e.}\xspace}
\def\eg{\emph{e.g.}}
\DeclareMathOperator*{\argmin}{arg\; min}

\cvprfinalcopy 

\def\cvprPaperID{****} 

\ifcvprfinal\fi
\begin{document}

\title{Tracking for Half an Hour}

\author{Ran~Tao, Efstratios~Gavves, Arnold~W.M.~Smeulders\\
QUVA Lab, 
University of Amsterdam\\
}

\maketitle
\thispagestyle{empty}

\begin{abstract}
Long-term tracking requires extreme stability to the multitude of model updates and robustness to the disappearance and loss of the target as such will inevitably happen. For motivation, we have taken 10 randomly selected OTB-sequences, doubled each by attaching a reversed version and repeated each double sequence 20 times. On most of these repetitive videos, the best current tracker performs worse on each loop. This illustrates the difference between optimization for short-term versus long-term tracking. In a long-term tracker a combined global and local search strategy is beneficial, allowing for recovery from failures and disappearance. Most importantly, the proposed tracker also employs cautious updating, guided by self-quality assessment. The proposed tracker is still among the best on the 20-sec OTB-videos while achieving state-of-the-art on the 100-sec UAV20L benchmark. 
On 10 new half-an-hour videos with city bicycling, sport games etc, the proposed tracker outperforms others by a large margin where the 2010 TLD tracker comes second.
\end{abstract}

\section{Introduction}
Fueled by the availability of standard datasets~\cite{kristan2015visual,wu2015object,smeulders2014visual}, tracking has made a large progress over the last few years. However, the videos in ALOV~\cite{smeulders2014visual} are about 10 seconds long on average and OTB~\cite{wu2015object} is about 20 seconds per video. The rationale for these relatively short episodes in ALOV, OTB and VOT was to select hard moments like a transition into a different illumination, abrupt motion, clutter, large shape change, sudden occlusions and some more factors of difficulty. It was implicitly perceived as when most hard moments can be solved, tracking of the episodes in between follows suit. Where this gives a good insight in why trackers fail, most surveillance, man-machine interactions, sport games, ego-documents or TV show videos are much longer. And, it appears that in real-life scenarios, when tracking for half an hour, other elements become important other than surviving the hardest short episodes.

In long videos, the above cited difficult episodes may also occur. In addition, tracking in long videos demonstrates challenges caused by the length of the video. In this paper we focus on these long term effects.

\section{Long-duration Tracking}\label{sec:long-term}
The best performing tracker on OTB and VOT~\cite{danelljan2017eco} performs much worse on half-an-hour videos, as we will show later in the experimental section.

Whether short or long, tracking starts from one observation of the target. After that, the tracker has to address a series of difficult tracking conditions, such as illumination variation, viewpoint change and deformation, in order to follow the target. The longer the video, the higher the chances one or more of these conditions will occur. Long-term tracking needs to be solidly robust to a large variety of circumstances including their combined effect.

Most current successful trackers~\cite{hare2011struck,HenriquesC0B15,DanelljanBMVC2014,danelljan2017eco,tao2016sint,bertinetto2016fully,nam2016learning} localize the target in a frame by searching over the location predicted from the previous frame. The underlying assumptions are that the prediction is accurate and the target moves slowly from a frame to the next one. When a tracking failure occurs in the previous frame, the first assumption breaks and the target is lost as the target is not in the local area analyzed by the tracker. We call this \textit{sampling drift}. Even when the previous prediction is correct, \ie, no tracking failure has occurred in the previous frame, sampling drift may still happen when the target moves fast or abruptly. And, when the video is long, the video will more likely contain video cuts. A video cut introduces a significant change in viewpoint and an abrupt jump from one frame to the next. In long-term tracking motion continuity cannot be assumed and sampling drift will occur if the tracker follows a local search strategy as the above cited trackers do.

In long videos, an intrinsic element is that the target may disappear from the camera view for a while and reappear again. Such a break in the trajectory is rarely present in handpicked short sequences, but they occur in almost every long video either by long occlusion or by out-of-frame. When the target reappears, it may enter the camera view from an arbitrary position. This provides another argument that motion continuity cannot be assumed. In addition, failure may occur when an unforeseen combination of these effects occurs. Therefore, a failure recovery mechanism is indispensable in long-term tracking. The disappearance and reappearance of the target, the video cuts and the length of the video itself all lead to higher chance of sampling drift, urging to go beyond local search.

Apart from sampling drift,
there is also an issue for long-term tracking with model updating. For long-term tracking, analysis of various trackers has indicated that model updates may eventually ruin the internal model when each model update is off by a small margin~\cite{smeulders2014visual}. These minor adverse updates are too small to have a decisive effect in short-term tracking, but in the long run, the updates will accumulate and eventually cause the target model to drift. This is due to the fact that model updates have no mechanism of knowing whether the update is exact.  
In long-term tracking model drift will occur, to be handled by a super robust model, or by an update strategy, cautiously capable of determining when to update.

\begin{figure}
    \centering 
    \includegraphics[width=0.8\linewidth]{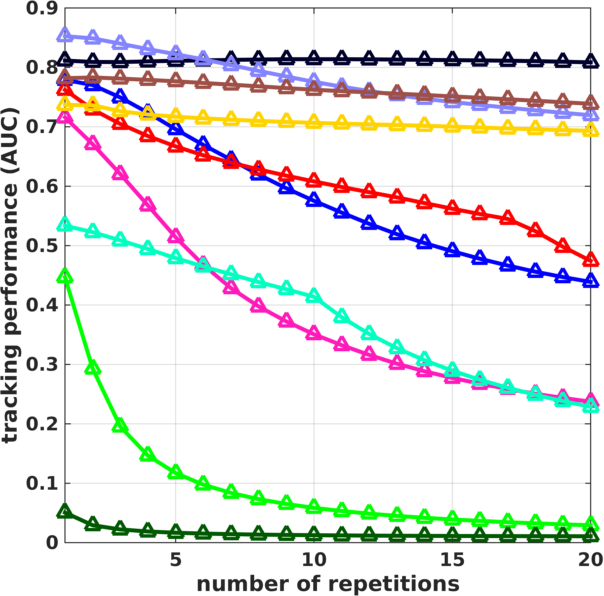} 
    \caption{For 10 randomly selected OTB sequences, we attach a reversed version of the video at the end by playing it backward. In this way, we create a sequence in which the target returns to the starting position. Then we repeat the created double sequence 20 times to make an increasingly long video without introducing any new difficulty. We evaluate ECO~\cite{danelljan2017eco}, the best performing tracker on OTB. The performance at the end of each loop steadily or rapidly decreases while looping through the same double sequence (except the one shown in black). Even when there is no new challenge, the best current tracker optimized for short sequences performs worse on each loop due to unstable model updates.}
    \label{fig:0_experiment}
    \vspace{-2mm}
\end{figure}

In Figure~\ref{fig:0_experiment} we conduct an experiment to demonstrate the length of the video itself brings challenges, even when there is no object disappearance and reappearance or video cut.

To summarize, in order to track the target for long duration, not only does the tracker have to address conventional difficult tracking conditions which short-term tracking focuses on, but also it has to be factors better at dealing with sampling drift and model drift. In this paper, we propose a long-duration tracker tackling the aforementioned issues.

\section{Related Work}
\subsection{Tracking by Detection}

Among the very many trackers, few pay attention to long-term tracking. We discuss~\cite{kalal2012tracking,supancic2013self,Pernici2014}.
TLD~\cite{kalal2012tracking} is a successful multi-component tracker, a classic. It accepts the principle of recovery by combining an optical flow tracker with a detector. The detector is updated cautiously in order to increase the robustness against model drift. The composite has a drawback in that the tracking and the detection will respond differently to different circumstances. A homogeneous model is to be preferred. In the evaluation on ALOV~\cite{smeulders2014visual}, TLD performed 4-th with many papers improving its performance on OTB and ALOV since. 
Also SPL~\cite{supancic2013self} follows the tracking by detection paradigm. To avoid model drift, the SVM-based detector is updated by selecting those frames which, when added to the training set, produce the lowest SVM objective. The repeated evaluation of the SVM objective to decide which frames to add, however, is computationally very expensive. 
Alien~\cite{Pernici2014} relies on oversampling of keypoints and RANSAC-based geometric matching to find the target. The tracker has a cautious mechanism for updating by verifying the quality of the geometric matching. Its use is, however, restricted to textured objects with simple rigid deformations. 
The proposed tracker is not a composite tracker like TLD, which integrates box predictions from multiple components. Unlike Alien, the proposed method is applicable to any type of target object. Different from the aforementioned trackers, in this paper, to avoid model drift, a deep self-evaluation module is proposed to explicitly evaluate the tracker's prediction and guide the model update accordingly.

MDNet~\cite{nam2016learning} is a recent successful tracking-by-detection tracker using deep learning. It employs a deep classification network as the detector. The last layer is specialized for each video, while the previous layers are shared across videos and pre-trained using external videos. MDNet achieves great performance on short-term tracking datasets OTB and VOT. However, it does not pay attention to long-term tracking. Its use of a risky update strategy and a local search scheme makes it not as robust against model drift and sampling drift. 
EBT~\cite{Zhu_2016_CVPR} goes beyond local search by generating instance-specific object proposals over the whole frame. An online learned and updated SVM classifier is used for proposal generation. As a consequence of the online learning, EBT has the risk of model drift even in the stage of proposal generation. Although EBT does not target at long-term tracking, the idea of going beyond local search would be beneficial for tracking in long videos. Similar to~\cite{Zhu_2016_CVPR}, this paper goes beyond local search. Differently, the proposed tracker employs a hybrid strategy that combines global search and local search. And the global search is performed periodically with a `time clock', which does not require any online learned model and hence has no additional risk of drift.


\subsection{Tracking by Correlation Filters}

A family of successful trackers are based on discriminative correlation filters (DCF).
Since the MOSSE tracker~\cite{BolmeBDL10}, many variants have been proposed.~\cite{Danelljan_2014_CVPR} uses multi-dimensional features.~\cite{HenriquesC0B15} proposes kernelized correlation filters. Robust scale estimation is incorporated~\cite{DanelljanBMVC2014}.~\cite{Danelljan2015,kiani2015correlation} address the boundary effects caused by the circular shift.~\cite{danelljan2016beyond} learns the filters in the continuous spatial domain, enabling the use of feature maps of different resolutions and allowing for sub-pixel localization.~\cite{danelljan2017eco} further improves~\cite{danelljan2016beyond} by addressing its over-fitting problem.~\cite{ma2015hierarchical,qi2016hedged,danelljan2016beyond,danelljan2017eco} integrate convolutional features with the DCF framework.
DCF trackers have shown great performance on tracking benchmarks of short videos~\cite{kristan2015visual,wu2015object}. However, relying on frequent, risky update schemes and local search, these trackers are not as robust against model drift and sampling drift.

LTCT~\cite{MaYZY15} is a DCF-style tracker paying attention to long-term tracking. It combines a DCF tracker with an online detector to re-detect the target in case of tracking failures. Failure detection is based on thresholding the confidence score of the DCF tracker. While the purpose is long-term tracking, the dataset used is OTB with an average length of 20 seconds.

While the performance of DCF trackers on short sequences like OTB and VOT is superior, we will demonstrate for the best performing DCF tracker~\cite{danelljan2017eco} the very limited success in long-term tracking. Even for LTCT which pays attention to long term, we will demonstrate likewise on half-an-hour videos.

\subsection{Tracking by Similarity Comparison}
Siamese trackers~\cite{bertinetto2016fully,tao2016sint} follow a tracking by similarity comparison strategy.
They simply search for the candidate most similar to the original image patch of the target given in the starting frame, using a run-time fixed but learned \textit{a priori} deep Siamese similarity function. Due to their no-updating nature, Siamese trackers are robust against model drift. However, this comes at the cost of not handling well confusions and drastic appearance changes. We draw inspiration from Siamese trackers, and employ a tracking by similarity comparison strategy. Different from Siamese trackers which employ a local search strategy, the proposed tracker uses a hybrid strategy combining global search and local search, and has the advantage of being robust against sampling drift. Unlike Siamese trackers which do not update at all, the proposed tracker employs a self-aware update strategy. As a result, the proposed tracker is better in handling confusing distractors and the significant appearance changes one would expect to occur in long videos, while still being robust against model drift.

\section{Method}

Inspired by the recent successful Siamese trackers~\cite{tao2016sint,bertinetto2016fully}, the 
proposed tracker employs a tracking by similarity comparison strategy. For an incoming frame, the tracker searches for the candidate most similar to the original image patch of the target given in the first frame. The similarity function is a deep two-branch Siamese network. 
To address sampling drift, the proposed tracker employs a novel search strategy that combines global search and local search. To address model drift, we propose a self-evaluation module that is capable of assessing the quality of the tracker's predictions, and a cautious model update strategy which updates the similarity function only when approved by the self-evaluation module. We describe the key elements of the tracker in detail in the following.

\subsection{Similarity Comparison}

The similarity function is formulated as a Siamese network, composed of two identical branches, each being a fully convolutional network~\cite{bertinetto2016fully}. One branch receives, as the query, the initial patch of the target in the first frame, denoted as $q$, and produces a 3D tensor representation $\phi(q)$. The other branch takes a frame or a cropped probe region, denoted as $p$, and produces $\phi(p)$. The similarity between the query $q$ and the candidates held in $p$, \ie, all translated windows in $p$ having the same size as $q$, is efficiently evaluated with a cross-correlation $f(q,p)=\phi(q)*\phi(p)$. The output of the Siamese network is a 2D similarity map $S=f(q,p)$. Each value on the map is the similarity between the query and the corresponding candidate.

\subsection{Hybrid Search}

The proposed hybrid search combines global search and local search. Global search searches for the target globally both in the spatial and scale spaces, preventing sampling drift. Local search only searches for the target locally around the predicted position in the previous frame, and over scales close to the previously estimated scale of the target. Local search is prone to sampling drift, but more efficient than global search. To take the advantage of both sides, we propose a hybrid strategy, performing global search once every $T$ frames and conducting local search on frames in between. The switch between global search and local search is decided with a ``time clock'', and it is not decided by reasoning about the tracker's prediction, as the latter would open a new door for cumulative error.

\textbf{Global search.} Global search is designed as a three-stage procedure for efficiency. 
In the first stage, the tracker searches at a single scale over the entire frame, and identifies $N$ potential locations $\{(u_i, v_i, w_{0}, h_{0})\}_{i=1}^{N}$ most similar to the initial patch of the target. $w_{0}$, $h_{0}$ are the width and height of the initial target in the first frame and $u$, $v$ are the coordinates of the box center. The aim of this stage is to have the target located in the local neighborhood of one of the $N$ locations.

In the second stage, the tracker searches locally around each of the $N$ locations over multiple scales $\{\sigma_{i}\}_{i=1}^{M}$, and selects the best box $\hat{b}=(\hat{u}, \hat{v}, \hat{w}, \hat{h})$. Specifically, $M\cdot N$ local probe regions $\{p_{ij}=(u_i, v_i, w_{0}\cdot{\sigma_{j}}\cdot{t}, h_{0}\cdot{\sigma_{j}}\cdot{t})|i=1...N,j=1...M\}$ are cropped from the frame. $t$ is a scaling factor. The similarity function takes the initial target $q$, resized to $l\times{l}$, and the probe regions $p_{ij}$, all resized to $tl\times{tl}$, and produces $M\cdot{N}$ 2D similarity maps. $\hat{b}$ is the box corresponding to the highest value on the similarity maps.

\begin{figure}
    \centering 
    \includegraphics[width=1\linewidth]{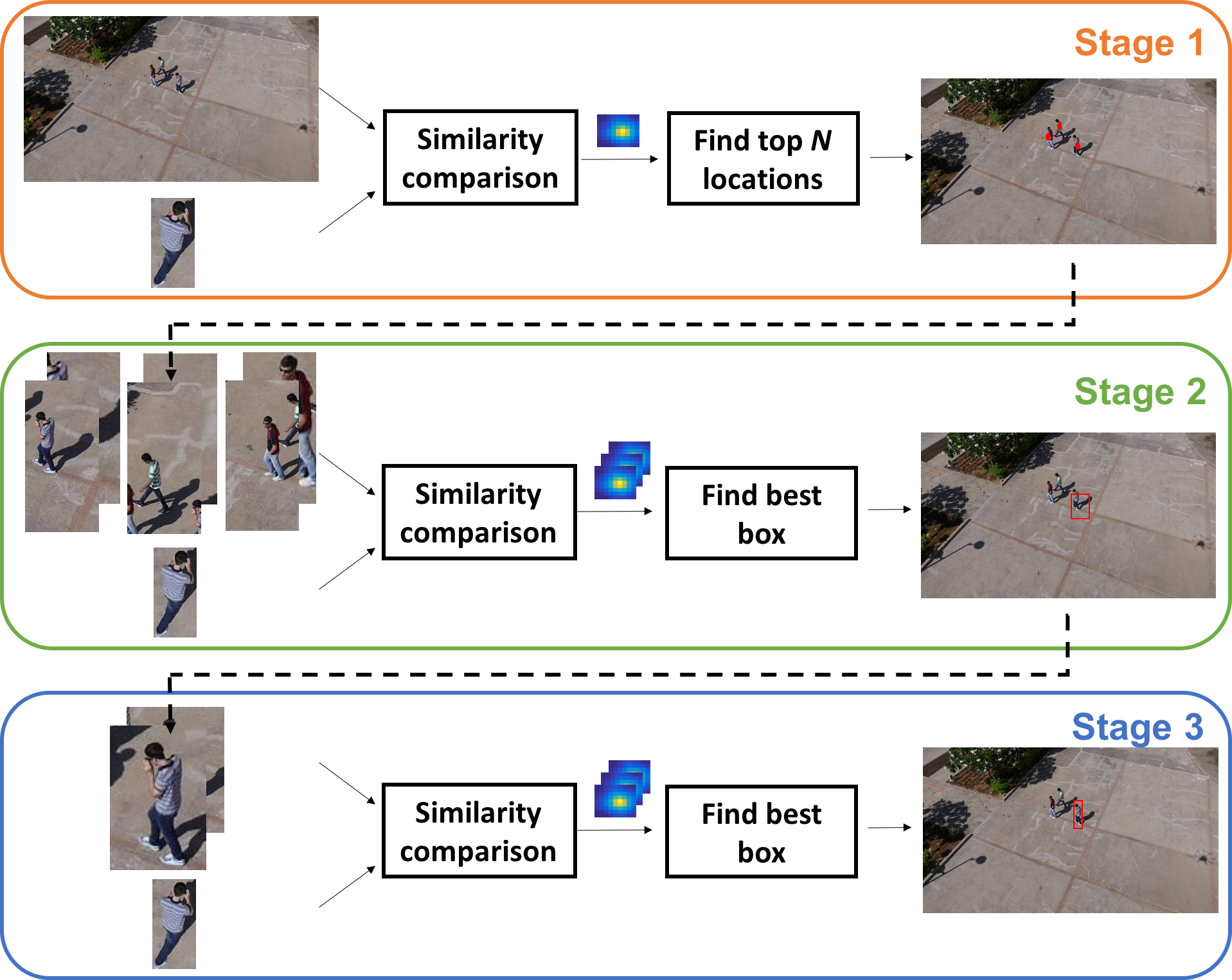} 
    \caption{The three-stage global search scheme. In stage 1, the tracker searches over the entire frame at a single scale, and identifies $N$ promising locations. In stage 2, around each of the $N$ locations, the tracker searches over multiple scales and returns the best candidate box. In stage 3, locally around the best box returned from stage 2, the tracker searches over multiple finer scales than in stage 2, using a larger input resolution than the previous two stages (not shown in the figure for clarity), to derive a better localization in both spatial and scale spaces.} 
    \label{fig:global_search}
\end{figure}

In the third stage, with a larger input resolution, the tracker searches around $\hat{b}$ over multiple finer scales $\{\tilde{\sigma_{i}}\}_{i=1}^{L}$ that span the scale interval of $\{\sigma_{i}\}_{i=1}^{M}$. The aim of the final stage is to derive a better localization in both spatial and scale spaces. Concretely, $L$ probe regions $\{\tilde{p_{j}}=(\hat{u}, \hat{v}, \hat{w}\cdot{\tilde{\sigma_{j}}}\cdot{t},\hat{h}\cdot{\tilde{\sigma_{j}}}\cdot{t})|j=1...L\}$ are sampled from the frame. The initial target $q$, resized to $\tilde{l}\times \tilde{l}$ where $\tilde{l}>l$, and the probe regions $\tilde{p_{j}}$, resized to $t\tilde{l}\times t\tilde{l}$, are input to the similarity function. The final prediction $\tilde{b}$ for the frame is determined by selecting the candidate corresponding to the largest value on the similarity maps. 
Figure~\ref{fig:global_search} illustrates the three-stage global search scheme.

\textbf{Local search.} Local search is similar to the third stage of global search. The tracker searches around the predicted location in the previous frame, with input resolution $l^{\prime}$, over multiple scales $\{\sigma^{\prime}\}$ close to the previously estimated scale of the target,  and returns the best box as the prediction.

\subsection{Self-aware Model Update}\label{sec:sa_upd}

\textbf{Self-evaluation module.} The objective of the self-evaluation module is to guide model update such that beneficial updates are kept and adverse updates are avoided as much as possible. We define an update to be beneficial if the tracker's predicted box is correct, \ie, the training data used to update the model are correct, and adverse if the prediction is wrong. Following this definition, we formulate self-evaluation as a binary classification problem, predicting whether the tracker's predicted box is correct.

\begin{figure}
    \centering 
    \includegraphics[width=0.9\linewidth]{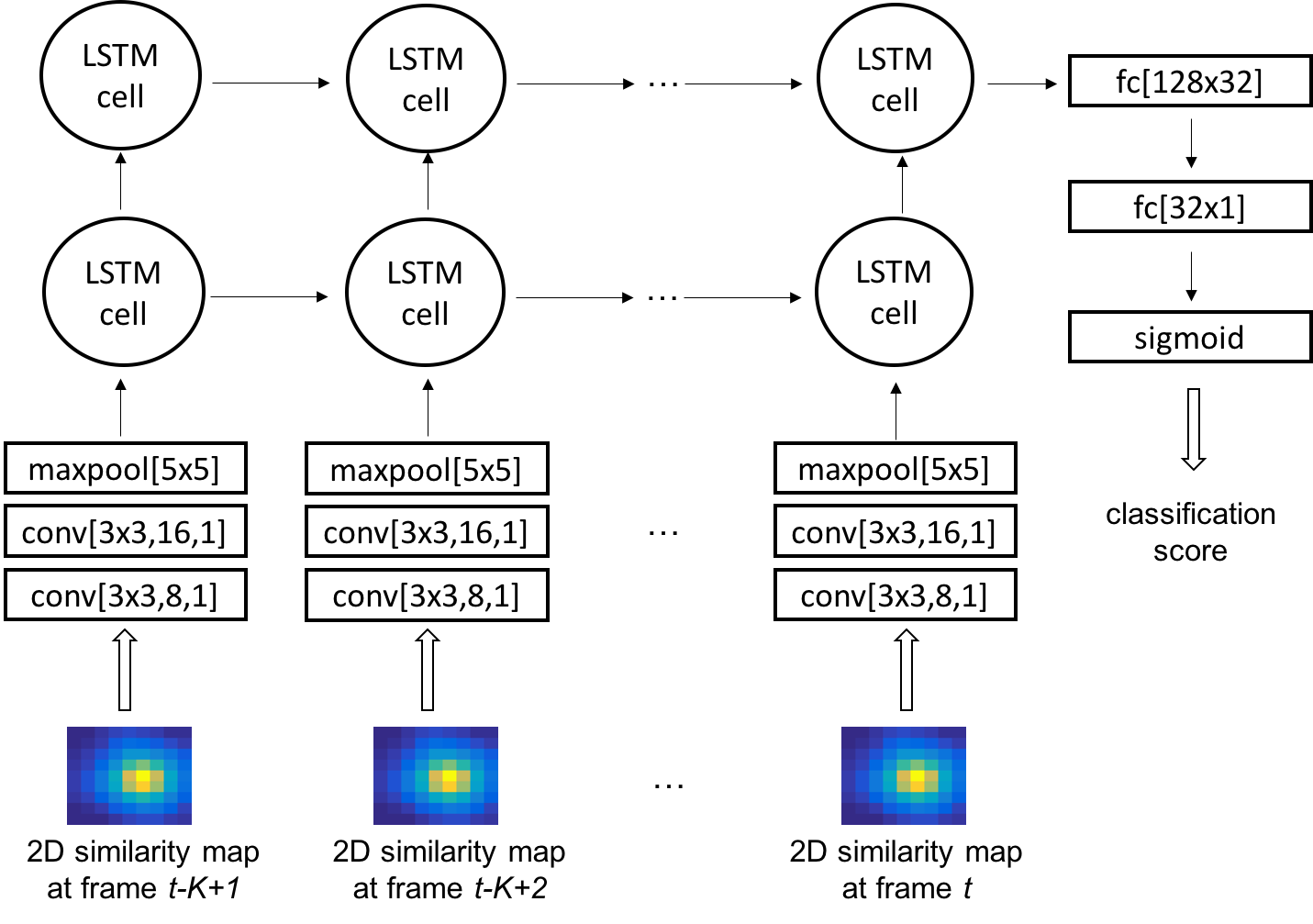} 
    \caption{The network architecture of the self-evaluation module. The module takes the similarity map from the current frame and the ones from previous $K-1$ frames as input, and classifies whether the tracker's prediction in the current frame is correct. The conv layers and the first fully connected layer are followed by ReLU~\cite{nair2010rectified}.}
    \label{fig:selfeval_module}
\end{figure}

An LSTM-based binary classifier is proposed. It conditions on the similarity map in the current frame and the ones in previous $K-1$ frames. For frames where global search is performed, the similarity map from the final stage is used. The similarity map includes information that is indicative of the quality of the tracker's prediction~\cite{choi2017visual}. Intuitively, when there is a sharp peak in the similarity map, it is likely that the peak corresponds to the true target. Similarity maps from history are incorporated to capture the temporal dynamics of the similarity distributions. The recurrent network architecture is shown in Figure~\ref{fig:selfeval_module}. The similarity maps are first encoded by a small convnet and the whole sequence is summarized by a two-layer LSTM network. The hidden representation from the last step is input to a two-layer multilayer perceptron to get the classification output. 

Given the training sequences $\mathcal{D}=\{(x_i, y_i)\}_{i=1}^{n}$ where $x_i$ is a sequence of similarity maps and $y_i\in \{0,1\}$ is the binary label, the classifier with parameters $\theta$ is trained by minimizing the binary cross entropy loss $\argmin_{\theta} -\frac{1}{n}\sum_{i=1}^{n}y_i\cdot \log g_i + (1-y_i)\cdot \log(1-g_i)$,
where $g_i$ is the classification output on $x_i$. \\

\textbf{Model update.} A cautious model update strategy is employed. Update is carried out only on frames where global search is performed 
and only when the self-evaluation module approves the quality of the tracker's prediction. As temporary sampling drift might occur on frames where local search is performed, the update takes place only after global search to disentangle model drift from sampling drift. In this way, the chance of taking adverse model updates is reduced. Furthermore, only the similarity function for stage 2 of the global search scheme is updated while the similarity functions for stage 1 and stage 3 are fixed. The aim of stage 1 is to include the target in the top $N$ retrieved locations, for which an offline learned similarity function is likely to be sufficient. And updating the model for stage 1 is extremely risky as once the model has drifted in stage 1 it will be impossible to find the target even when models in stage 2 and 3 are perfect. Hence, we do not update the similarity function for stage 1. The similarity function for stage 3 is also frozen, since the purpose of stage 3 is to refine the localization, similar to the box regression employed in~\cite{tao2016sint,nam2016learning}, for which an offline learned similarity is sufficient. In stage 2, the task is to find the true target from a set of candidates which are all similar to the initial target. Therefore, in stage 2, the tracker needs to deal with confusing distractors, for which online adapting the model would be beneficial.

The similarity function for stage 2 is updated using the training pairs formed as follows. We pair the initial target patch from the first frame, $q$, and the final predicted box on the current frame as the positive training sample. And for negative samples, we make pairs between the initial target patch and all the candidate patches held in the other $M\cdot{(N-1)}$ probe regions considered in stage 2 as long as they do not overlap with the final prediction. These are hard negative samples which serve the purpose of adapting the similarity function to handle confusions. 

With the training data $\mathcal{P}=\{(q, b_i, y_i)\}_{i=1}^{m}$ where $y_i\in\{0,1\}$, the Siamese network of the similarity function with parameters $\theta_{s}$ is updated by minimizing the binary cross entropy loss, \ie, $\argmin_{\theta_{s}}-\frac{1}{m}\sum_{i=1}^{m}y_i\cdot \log s_i + (1-y_i)\cdot \log(1-s_i)$, where $s_i$ is the similarity between $q$ and $b_i$, normalized to $[0,1]$.

\section{Experiments}
\subsection{Implementation Details}

\textbf{Network architecture.} The Siamese network is composed of two identical branches. The architecture of the branch network is the same as the VGG-16 network~\cite{simonyan2015very}, till \texttt{relu4\_3}. It consists of 10 convolution layers and 3 2-by-2 max pooling layers. Convolution layers are followed by ReLU~\cite{nair2010rectified}.

\textbf{Hybrid search.} The first stage of global search identifies $N(=10)$ candidate locations most likely to contain the target in the local neighborhood. In the second stage, the tracker search locally around the $N$ locations ($t=2$), over $M(=9)$ scales, $\{\sigma\}=2^{\{-2:0.5:2\}}$, and returns the best box. In the first two stages, the query patch of the initial target is resized to $32\times32$, \ie, $l=32$. In the first stage, the whole frame is resized accordingly and in the second stage, the probe regions are resized to $tl\times{tl}=64\times64$. The third stage refines the box returned in the previous stage by searching over $L(=11)$ finer scales, $\{\tilde{\sigma}\}=2^{\{-0.4:0.08:0.4\}}$ and using a larger input resolution, $\tilde{l}=64$. Local search is similar to the third stage of global search. The tracker searches locally ($t=2$) around the previous estimated location over 5 scales that are close to the previously estimated scale, $\{\sigma^{\prime}\}=\{0.9509,0.9751,1,1.0255,1.0517\}$, following~\cite{bertinetto2016fully}. The input resolution for local search is $l^{\prime}(=64)$.

\textbf{Self-aware model update.} 
The data for training the self-evaluation module are generated from ALOV~\cite{smeulders2014visual} excluding the ones which also appear in OTB~\cite{wu2015object}, by running a variant of the proposed tracker. The variant runs global search on very frame and does not update the model online. $^5/_6$ of the videos are used for training and the rest for validation. The sequence length is $K(=10)$. The binary label is determined based on the intersection-over-union (IoU) between the predicted box and the groundtruth. A training sample is deemed positive if the IoU is over 0.5, and deemed negative otherwise. To bias the self-evaluation module towards being conservative and make the training stable, samples are assigned different weights during training. Specifically, the weights are 1, 0.05 and 0.3 for samples with $IoU<0.3$, $IoU\in[0.3,0.5]$ and $IoU>0.5$ respectively. The weights are determined using the validation set. 
The self-evaluation module is trained offline. 
When a model update is permitted during online tracking, the Siamese network of the similarity function is updated using SGD with momentum for 10 iterations, with the learning rate and momentum being 0.01 and 0.9.

\subsection{Experiments on Long-term Tracking}

\subsubsection{Datasets and Evaluation Metric}

\textbf{UAV20L.} 
UAV20L contains 20 videos captured from low-altitude unmanned aerial vehicles. It was recently proposed in~\cite{mueller2016benchmark} for long-term tracking evaluation. Compared to OTB~\cite{wu2015object} and VOT~\cite{kristan2015visual}, 
the videos in UAV20L are longer, with an average length of about 100 seconds. We evaluate on UAV20L as it has the longest videos among existing tracking benchmarks, although 100 seconds is not very long.

\begin{figure*}
    \centering 
    \includegraphics[width=1.0\linewidth]{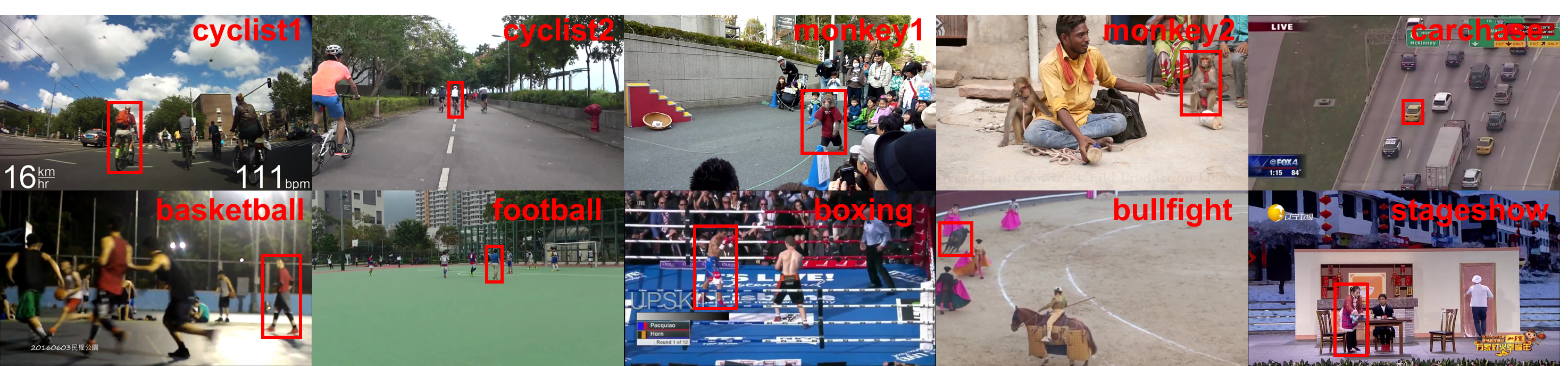} 
    \caption{Example frames from the 10 very long sequences.}
    \label{fig:ytlong_example_frames}
\end{figure*}

\textbf{YoutubeLong.} 
To evaluate on much longer videos than a few minutes, 
we gathered 10 very long videos from YouTube. The average video length is 25 minutes. 9 videos out of 10 are longer than 20 minutes with the longest being over 33 minutes. The videos are annotated in a sparse manner, one annotation every 100 frames. When the target is visible, a bounding box is annotated while frames where the target is invisible are marked as \textit{absent}. Figure~\ref{fig:ytlong_example_frames} shows an example frame for each video. In addition to being long, these sequences feature all sorts of challenging factors~\cite{wu2015object}, such as illumination variation, viewpoint change, non-rigid deformation, background clutter, confusion, abrupt motion and occlusion. Moreover, 
in these long videos, the target is absent for a significant portion of time. On average, the target is not present on over $16\%$ of the annotated frames, whereas in UAV20L it is only about $4\%$.\\

\textbf{Evaluation metric.} We employ the AUC metric used in OTB~\cite{wu2015object} and UAV20L~\cite{mueller2016benchmark} with a modification. The modification is made to evaluate the trackers better on videos where the target might be absent for a while. When the target is visible in the frame, the IoU between the predicted box and the ground-truth is computed. When the target is not visible, a tracker's prediction is considered to have 100\% IoU if the tracker explicitly predicts \textit{absence}. Any predicted box on the frame where the target is not visible gets 0 IoU. A frame is declared to be a success if the IoU is larger than a threshold, and the percentage of successfully tracked frames is calculated. A curve is created by varying the threshold and AUC is the area under the curve. 
We denote the modified AUC metric still as \textit{AUC} for convenience.

\subsubsection{Evaluation of Hybrid Search}

We first compare global search and local search for the task of tracking for long duration by running two variants, one performing global search on every frame and the other conducting local search on every frame. Online model update is disabled in both to ensure a fair comparison. 
The results are shown in Table~\ref{tab:global_vs_local}. Global search works clearly better than local search on both datasets, and the performance gap is larger on the YoutubeLong dataset. 
Local search relies on strong assumptions that the prediction in the previous frame is accurate and the target moves slowly from one frame to the next. These assumptions easily break in long videos where the target might disappear and reappear multiple times. Consequently, local search suffer from sampling drift. On the other hand, global search is free of sampling drift as it does not depend on the previous prediction and does not make any assumption on the motion pattern of the target. We conclude that global search is advantageous when tracking for long duration. 

\begin{table}
    \renewcommand{\arraystretch}{1.0}
    \centering
    \scalebox{0.85} {
    \setlength{\tabcolsep}{6pt}
    \begin{tabular}{lcc}
        \toprule
        & \emph{UAV20L} & \emph{YoutubeLong}  \\
        \midrule
          \textit{Local Search} & 39.8 & 23.3 \\
          \textit{Global Search} & 49.4 & 37.6 \\
        \bottomrule
    \end{tabular}
    }
    \caption{Comparison between global search and local search on UAV20L and YoutubeLong, measured in AUC (\%). Global search is advantageous when tracking for long duration where the target might disappear and reappear.}
    \label{tab:global_vs_local}
\end{table}

\begin{figure}
    \centering 
    \includegraphics[width=0.9\linewidth]{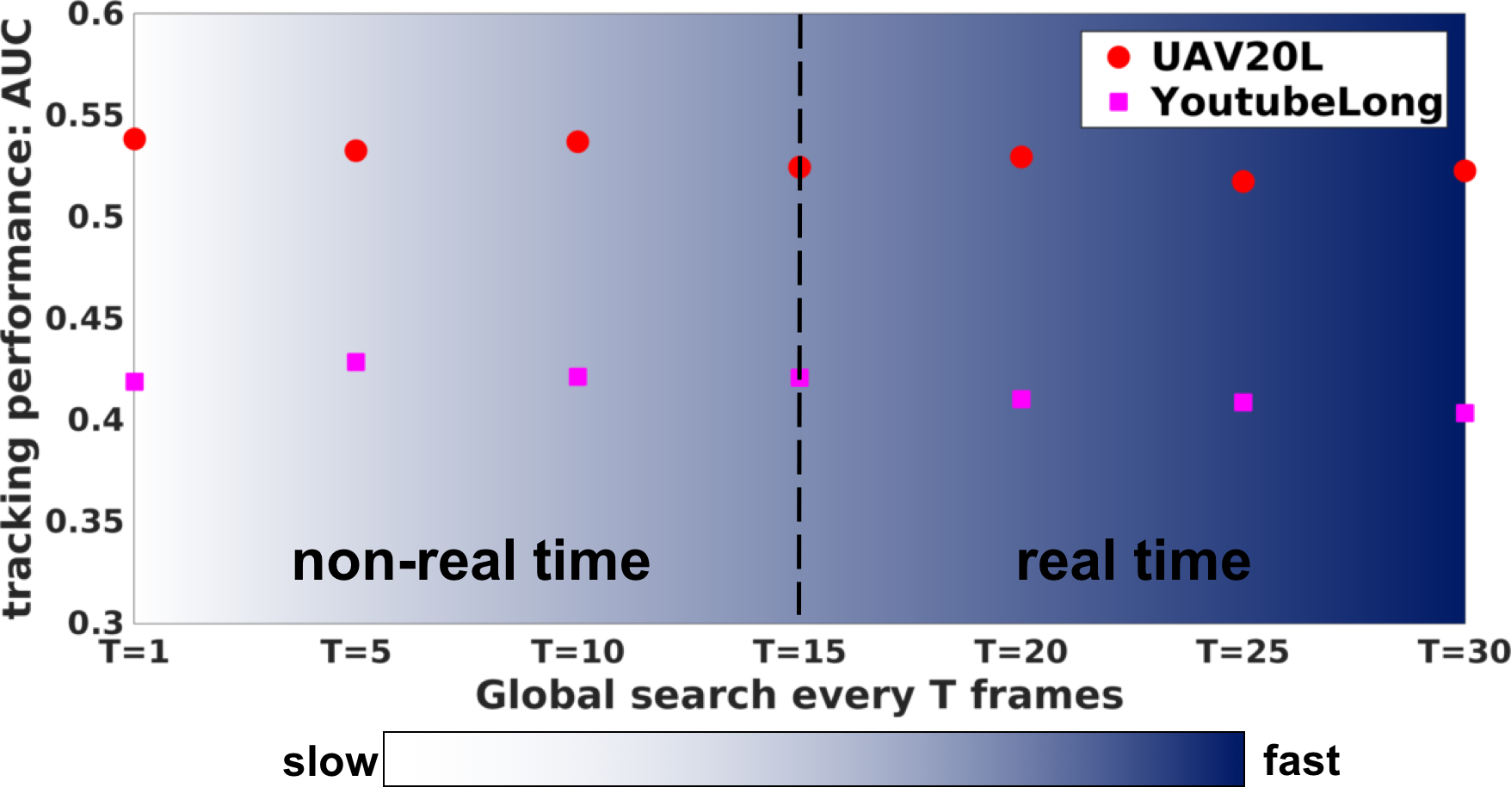} 
    \caption{The impact of the frequency of applying global search on tracking speed and accuracy. The proposed hybrid search applies global search once every $T$ frames. The tracker reaches real-time efficiency when global search is applied every 15 frames. As $T$ varies, there are mild changes in tracking accuracy.}
    \label{fig:hybrid_search_T}
\end{figure}

Next we evaluate the hybrid search strategy, and quantify the impact of the frequency of applying global search. Here self-aware model update is included. Results are shown in Figure~\ref{fig:hybrid_search_T}. As $T$ increases, \ie, the frequency of performing global search decreases, the tracker gets more efficient. The tracker reaches real-time performance when global search is performed every 15 frames. As $T$ varies, there are mild changes in tracking accuracy. We conclude that the proposed hybrid search is effective. It enjoys the advantage of global search, \ie, robust against sampling drift, and meanwhile enables real-time efficiency.

\subsubsection{Evaluation of Self-aware Model Update}

In this experiment, we evaluate the effectiveness of the proposed self-aware model update. To that end, three baselines are constructed for comparison. All the three baselines follow the same tracking procedure as the proposed tracker where the only difference lies in how to determine whether an update should be performed. We denote the three baselines as ``no-upd'', ``blind-upd'' and ``sim-upd''. The baseline ``no-upd'' does not update its similarity function at all. The baseline ``blind-upd'' simply updates every time. The baseline ``sim-upd'' conducts model update when the similarity score of the predicted target box is above a certain threshold ($0.5$ in this experiment). Assessing the tracking status based on the score of the tracker's prediction and making decisions by thresholding the score has been used before in the literature, \eg,~\cite{MaYZY15,nam2016learning}. In this experiment, global search is conducted on every frame. Table~\ref{tab:sa_upd} lists the results. 
On UAV20L, the baseline ``blind-upd'' improves over ``no-upd'' whereas on YoutubeLong ``blind-upd'' has a large drop in performance. 
Because of the aerial nature of the UAV20L dataset, the appearance variation is mild and the motion is quite smooth. Also in UAV20L the target absence is not frequent. However, the videos in YoutubeLong are much longer, more wild and with more frequent target absence. Consequently, the tracker encounters more tracking failures on YoutubeLong, and without any cautious mechanism, takes more adverse model updates. Therefore, for ``blind-upd'', the model gets drifted on half-an-hour videos. 
``sim-upd'' and the proposed ``selfaware-upd'' are more robust against model drift than ``blind-upd'', thanks to their cautious update schemes. 
On both datasets, the proposed self-aware model update achieves the best performance. We conclude the proposed self-aware model update is effective in handling model drift.

\begin{table}
    \renewcommand{\arraystretch}{1.0}
    \centering
    \scalebox{0.85} {
    \setlength{\tabcolsep}{6pt}
    \begin{tabular}{lcc}
        \toprule
        & \emph{UAV20L} & \emph{YoutubeLong}  \\
        \midrule
          \textit{no-upd} & 49.4 & 37.6 \\
          \textit{blind-upd} & 50.4 & 25.3 \\
          \textit{sim-upd} & 51.4 & 40.4 \\
          \textit{selfaware-upd} & 53.9 & 41.9 \\
        \bottomrule
    \end{tabular}
    }
    \vspace{-2mm}
    \caption{Comparison between the proposed self-aware model update and three baselines, measured in AUC (\%). ``no-upd'' does not update the model. ``blind-upd'' updates every time. ``sim-upd'' updates the model when the similarity score of the predicted box is above the threshold (0.5). The proposed ``selfaware-upd'' is effective in handling model drift, achieving the best performance.}
    \label{tab:sa_upd}
\end{table}

\subsubsection{State-of-the-art Comparison}

We compare the proposed approach with state-of-the-art methods, including ECO-DEEP~\cite{danelljan2017eco}, TLD~\cite{kalal2012tracking}, LTCT~\cite{MaYZY15}, SPL~\cite{supancic2013self}, SRDCF~\cite{Danelljan2015} and MUSTer~\cite{Hong2015}. 
ECO-DEEP does not aim for long-term tracking. We evaluate it here as it is the best performing tracker on short-term tracking benchmarks for the moment.
TLD is a classic tracker, paying attention to long-term tracking. 
LTCT has a re-detection component, aiming for long-term tracking. 
SPL also pays attention to long-term tracking. However, SPL is very slow. The implementation provided by the authors runs at about $0.05$ frames per second. We were only able to evaluate it on the relatively small UAV20L, on which the experiment took about 2 weeks. It would take about 3 months to evaluate SPL on the much longer videos in YoutubeLong. 
SRDCF and MUSTer are the two best performing trackers on UAV20L, according to the evaluation in~\cite{mueller2016benchmark} when proposing the dataset, therefore, we also report their results on UAV20L for comparison. 

The results are summarized in Table~\ref{tab:state_of_the_art}. 
ECO-DEEP, the best performing tracker on short-term benchmarks, OTB and VOT, still works well on UAV20L which is about 100 seconds per video. However, on the half-an-hour videos in YoutubeLong, ECO-DEEP works poorly. Due to its use of risky model update scheme and local search, ECO-DEEP is not suited for the long videos with object absence. TLD works reasonably well even on the very long YoutubeLong, as it has a failure recovery mechanism by combing an optical flow tracker with a detector. LTCT has no success on YoutubeLong, although it pays attention to long-term tracking. The proposed tracker achieves the best performance on both datasets, outperforming others by a large margin. On UAV20L, the proposed tracker achieves $52.4\%$ in AUC, about $10\%$ better than the second, setting a new state-of-the-art. On the newly proposed long videos, the advantage of the proposed tracker is even more clear.

In addition, we also evaluate the trackers' performance on the last 20 seconds of each video in YoutubeLong, to show the trackers' capabilities of following the target till the end. The results are listed in the last column of Table~\ref{tab:state_of_the_art}. Only the proposed method and TLD are capable of following the target to the end, and the proposed method is better. 
\vspace{-2mm}

\begin{table}
    \renewcommand{\arraystretch}{1.0}
    \centering
    \scalebox{0.75} {
    \setlength{\tabcolsep}{6pt}
    \begin{tabular}{lccc}
        \toprule
        & \emph{UAV20L} & \emph{YoutubeLong} & \emph{YoutubeLong}  \\
        & \emph{(100 sec)}& \emph{(half-an-hour)} &\emph{(last 20 sec)} \\
        \midrule
          ECO-DEEP~\cite{danelljan2017eco} & 42.7 & 7.1 & 1.4 \\
          TLD~\cite{kalal2012tracking} & 22.8 & 22.4 & 20.2\\
          LTCT~\cite{MaYZY15} & 25.5 & 2.2 & 0.2\\
          SPL~\cite{supancic2013self} & 35.6 &  \textemdash & \textemdash \\
          SRDCF~\cite{Danelljan2015} & 34.3 & \textemdash & \textemdash\\
          MUSTer~\cite{Hong2015} & 32.9 & \textemdash & \textemdash\\
          This paper & \textbf{52.4} & \textbf{42.1} & \textbf{39.5}\\
        \bottomrule
    \end{tabular}
    }
    \vspace{-2mm}
    \caption{State-of-the-art comparison, measured in AUC (\%). On UAV20L all the trackers show some success. However, on the much longer YoutubeLong which also contains more target absence, only the proposed method ($T=15$ here) and TLD show success, capable of following the target to the end. On both datasets, the proposed method outperforms others by a large margin.}
    \label{tab:state_of_the_art}
    \vspace{-3mm}
\end{table}

\subsubsection{Experiments on Repetitive Videos}

\begin{figure}
    \centering 
    \includegraphics[width=1\linewidth]{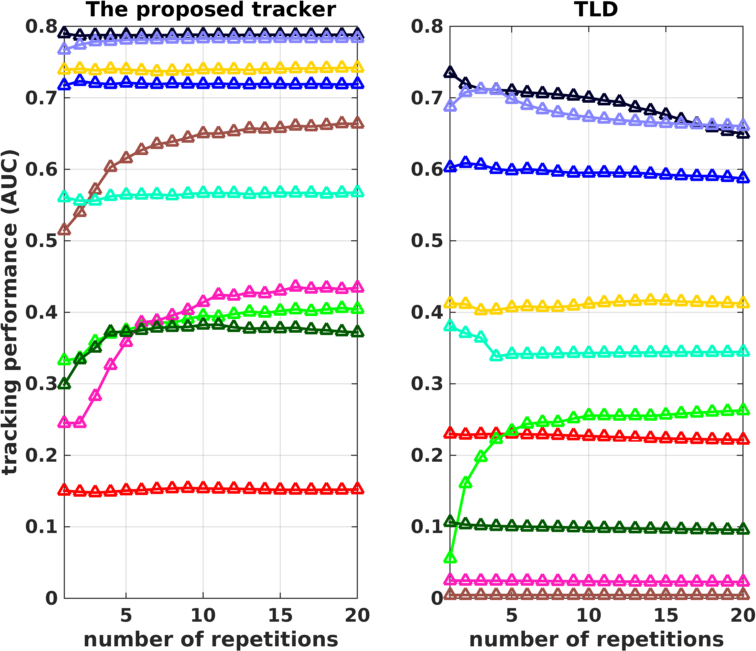} 
    \caption{Evaluate the proposed method and TLD~\cite{kalal2012tracking} on 10 repetitive videos, see Figure~\ref{fig:0_experiment} for details. Here, video length has no negative impact on the proposed tracker, it even benefits from the increasing length in 4 out of 10 videos. TLD is also very stable with a gain in 1 and loss in 2 out of 10. Compared to modern trackers like ECO~\cite{danelljan2017eco} (see Figure~\ref{fig:0_experiment}), the proposed tracker and the 2010 TLD tracker are superior. Among these videos, the proposed tracker outperforms TLD by a large margin.}
    \label{fig:0_experiment_ours_TLD}
\end{figure}

Now we come back to the 10 repetitive videos described in Section~\ref{sec:long-term}. We evaluate the proposed tracker and TLD~\cite{kalal2012tracking} on these videos. The results are shown in Figure~\ref{fig:0_experiment_ours_TLD}. The proposed method is very stable as the video length increases. On all the 10 videos, there is no decrease in performance as the length increases. Moreover, on 4 videos, there is a clear increase in performance, which is due to the beneficial model updates. Similarly, TLD is also very stable. Compared to modern trackers like ECO~\cite{danelljan2017eco} (see Figure~\ref{fig:0_experiment}), the proposed tracker and TLD are superior. Among these videos, the proposed tracker outperforms TLD by a large margin.

\subsection{Experiments on Short-term Tracking}

We also evaluate the proposed tracker on the short videos of OTB~\cite{wu2015object}. In this experiment, we use the standard AUC metric used by the benchmark. In Table~\ref{tab:otb100} we compile an overview of the performance of the state-of-the-art trackers. When applied to short-term tracking, the proposed tracker is comparable to the state-of-the-art trackers which focus on short-term tracking, although the proposed tracker focuses on long-term tracking.

\begin{table}
    \renewcommand{\arraystretch}{1.0}
    \centering
    \scalebox{0.7} {
    \setlength{\tabcolsep}{6pt}
    \begin{tabular}{lcccc}
        \toprule
        Method & \emph{OTB100~\cite{wu2015object}} &  &\multicolumn{1}{l}{Method} & \emph{OTB100~\cite{wu2015object}}  \\
        \midrule

        TLD~\cite{kalal2012tracking} (2010) & 40.6 & &\multicolumn{1}{l}{SINT~\cite{tao2016sint} (2016)} & 59.2 \\

        LTCT~\cite{MaYZY15} (2015) & 56.2 &  &\multicolumn{1}{l}{MDNet~\cite{nam2016learning} (2016)} & 67.8 \\

        MUSTer~\cite{Hong2015} (2015) & 57.2 & & \multicolumn{1}{l}{ECO~\cite{danelljan2017eco} (2017)} & 69.1 \\

        SRDCF~\cite{Danelljan2015} (2015) & 59.8 & &\multicolumn{1}{l}{CSRDCF~\cite{lukevzivc2017discriminative} (2017)} & 58.7 \\

        Staple~\cite{Bertinetto_2016_CVPR} (2016) & 58.1 & &\multicolumn{1}{l}{CFNet~\cite{valmadre2017end} (2017)} & 58.6 \\

        SiamFC~\cite{bertinetto2016fully} (2016) & 58.2 && \multicolumn{1}{l}{This paper} & 59.8 \\

        \bottomrule
    \end{tabular}
    }
    \vspace{-2mm}
    \caption{State-of-the-art comparison on the short-term tracking benchmark OTB~\cite{wu2015object}, measured in AUC\%. The proposed tracker is comparable to the state-of-the-art trackers that focus on short-term tracking, although the proposed tracker focuses on long-term tracking.}
    \label{tab:otb100}
\end{table}

\section{Conclusion}
This paper considers long-term tracking. Surprisingly, tracking for half an hour is very different from tracking the short videos in OTB, ALOV and other datasets, which have boosted the development of trackers over the last five years so eminently. 

In an experiment to motivate the research we consider 10 randomly selected videos from OTB. Each copy is expanded by a copy in reverse oder to arrive at the same position in the field of view. Then, 20 copies of these (forward, backward) pairs are glued together. The best current tracker on OTB, ECO~\cite{danelljan2017eco}, was selected to run on these 10 repetitive videos. It was noted that on most of the videos the tracker's performance was worse after each loop. This was expected as short-term trackers do not require stability of the model. The experiment (Figure~\ref{fig:0_experiment}) shows model deteriorates after each loop. 

In reaction to these observations we present a tracker specialising in long-term tracking. The tracker employs a global and local search strategy which allows recovery from an occasional failure and the occasional disappearance from the field of view. These events will always happen in a long video. In addition, a self-evaluation module is proposed, capable of assessing the quality of the tracker's predictions and cautiously guilding
the model update to be robust against model drift. 

We demonstrate that these two qualities of the proposed tracker are crucial to follow the target persistently even for half an hour. If the video is constantly streaming, \ie, the video is infinitely long, the conservative version of the proposed tracker without updating is guaranteed not to derail and still deliver a good performance on the 10 new realistic long Youtube videos we have collected and annotated from city adventures, sport games and alike. The advanced version with caution in updating does not derail for this 10 half-an-hour videos either. The tracker still is sufficiently good on OTB and much better than existing trackers on long and very long videos, where the solid 2010 TLD tracker~\cite{kalal2012tracking} now comes second again but by a wide margin.

{\small
\bibliographystyle{ieee}
\bibliography{main_ref}

\begin{thebibliography}{10}\itemsep=-1pt

\bibitem{Bertinetto_2016_CVPR}
L.~Bertinetto, J.~Valmadre, S.~Golodetz, O.~Miksik, and P.~H.~S. Torr.
\newblock Staple: Complementary learners for real-time tracking.
\newblock In {\em CVPR}, June 2016.

\bibitem{bertinetto2016fully}
L.~Bertinetto, J.~Valmadre, J.~F. Henriques, A.~Vedaldi, and P.~H. Torr.
\newblock Fully-convolutional siamese networks for object tracking.
\newblock In {\em ECCV VOT workshop}, 2016.

\bibitem{BolmeBDL10}
D.~S. Bolme, J.~R. Beveridge, B.~A. Draper, and Y.~M. Lui.
\newblock Visual object tracking using adaptive correlation filters.
\newblock In {\em CVPR}, 2010.

\bibitem{choi2017visual}
J.~Choi, J.~Kwon, and K.~M. Lee.
\newblock Visual tracking by reinforced decision making.
\newblock {\em arXiv preprint arXiv:1702.06291}, 2017.

\bibitem{danelljan2017eco}
M.~Danelljan, G.~Bhat, F.~S. Khan, and M.~Felsberg.
\newblock Eco: Efficient convolution operators for tracking.
\newblock In {\em CVPR}, 2017.

\bibitem{DanelljanBMVC2014}
M.~Danelljan, G.~Hager, F.~Shahbaz~Khan, and M.~Felsberg.
\newblock Accurate scale estimation for robust visual tracking.
\newblock In {\em BMVC}, 2014.

\bibitem{Danelljan2015}
M.~Danelljan, G.~Häger, F.~Khan, and M.~Felsberg.
\newblock Learning spatially regularized correlation filters for visual
  tracking.
\newblock In {\em ICCV}, 2015.

\bibitem{danelljan2016beyond}
M.~Danelljan, A.~Robinson, F.~S. Khan, and M.~Felsberg.
\newblock Beyond correlation filters: learning continuous convolution operators
  for visual tracking.
\newblock In {\em ECCV}, 2016.

\bibitem{Danelljan_2014_CVPR}
M.~Danelljan, F.~Shahbaz~Khan, M.~Felsberg, and J.~van~de Weijer.
\newblock Adaptive color attributes for real-time visual tracking.
\newblock In {\em CVPR}, 2014.

\bibitem{hare2011struck}
S.~Hare, A.~Saffari, and P.~H. Torr.
\newblock Struck: Structured output tracking with kernels.
\newblock In {\em ICCV}, 2011.

\bibitem{HenriquesC0B15}
J.~F. Henriques, R.~Caseiro, P.~Martins, and J.~Batista.
\newblock High-speed tracking with kernelized correlation filters.
\newblock {\em TPAMI}, 2015.

\bibitem{Hong2015}
Z.~Hong, Z.~Chen, C.~Wang, X.~Mei, D.~Prokhorov, and D.~Tao.
\newblock Multi-store tracker (muster): A cognitive psychology inspired
  approach to object tracking.
\newblock In {\em CVPR}, 2015.

\bibitem{kalal2012tracking}
Z.~Kalal, K.~Mikolajczyk, and J.~Matas.
\newblock Tracking-learning-detection.
\newblock {\em TPAMI}, 34(7):1409--1422, 2010.

\bibitem{kiani2015correlation}
H.~Kiani~Galoogahi, T.~Sim, and S.~Lucey.
\newblock Correlation filters with limited boundaries.
\newblock In {\em CVPR}, 2015.

\bibitem{kristan2015visual}
M.~Kristan, J.~Matas, A.~Leonardis, M.~Felsberg, L.~Cehovin, G.~Fernandez,
  T.~Vojir, G.~Hager, G.~Nebehay, and R.~Pflugfelder.
\newblock The visual object tracking vot2015 challenge results.
\newblock In {\em ICCV VOT workshop}, 2015.

\bibitem{lukevzivc2017discriminative}
A.~Luke{\v{z}}i{\v{c}}, T.~Voj{\'\i}{\v{r}}, L.~{\v{C}}ehovin, J.~Matas, and
  M.~Kristan.
\newblock Discriminative correlation filter with channel and spatial
  reliability.
\newblock In {\em CVPR}, 2017.

\bibitem{ma2015hierarchical}
C.~Ma, J.-B. Huang, X.~Yang, and M.-H. Yang.
\newblock Hierarchical convolutional features for visual tracking.
\newblock In {\em ICCV}, 2015.

\bibitem{MaYZY15}
C.~Ma, X.~Yang, C.~Zhang, and M.-H. Yang.
\newblock Long-term correlation tracking.
\newblock In {\em CVPR}, 2015.

\bibitem{mueller2016benchmark}
M.~Mueller, N.~Smith, and B.~Ghanem.
\newblock A benchmark and simulator for uav tracking.
\newblock In {\em ECCV}, 2016.

\bibitem{nair2010rectified}
V.~Nair and G.~E. Hinton.
\newblock Rectified linear units improve restricted boltzmann machines.
\newblock In {\em ICML}, 2010.

\bibitem{nam2016learning}
H.~Nam and B.~Han.
\newblock Learning multi-domain convolutional neural networks for visual
  tracking.
\newblock In {\em CVPR}, 2016.

\bibitem{Pernici2014}
F.~Pernici and A.~D. Bimbo.
\newblock Object tracking by oversampling local features.
\newblock {\em TPAMI}, 2014.

\bibitem{qi2016hedged}
Y.~Qi, S.~Zhang, L.~Qin, H.~Yao, Q.~Huang, J.~Lim, and M.-H. Yang.
\newblock Hedged deep tracking.
\newblock In {\em CVPR}, 2016.

\bibitem{simonyan2015very}
K.~Simonyan and A.~Zisserman.
\newblock Very deep convolutional networks for large-scale image recognition.
\newblock In {\em ICLR}, 2015.

\bibitem{smeulders2014visual}
A.~W.~M. Smeulders, D.~M. Chu, R.~Cucchiara, S.~Calderara, A.~Dehghan, and
  M.~Shah.
\newblock Visual tracking: an experimental survey.
\newblock {\em TPAMI}, 36(7):1442--1468, 2014.

\bibitem{supancic2013self}
J.~S. Supancic and D.~Ramanan.
\newblock Self-paced learning for long-term tracking.
\newblock In {\em CVPR}, 2013.

\bibitem{tao2016sint}
R.~Tao, E.~Gavves, and A.~W.~M. Smeulders.
\newblock Siamese instance search for tracking.
\newblock In {\em CVPR}, 2016.

\bibitem{valmadre2017end}
J.~Valmadre, L.~Bertinetto, J.~F. Henriques, A.~Vedaldi, and P.~H. Torr.
\newblock End-to-end representation learning for correlation filter based
  tracking.
\newblock 2017.

\bibitem{wu2015object}
Y.~Wu, J.~Lim, and M.-H. Yang.
\newblock Object tracking benchmark.
\newblock {\em TPAMI}, 37(9):1834--1848, 2015.

\bibitem{Zhu_2016_CVPR}
G.~Zhu, F.~Porikli, and H.~Li.
\newblock Beyond local search: Tracking objects everywhere with
  instance-specific proposals.
\newblock In {\em CVPR}, 2016.

\end{thebibliography}
}

\end{document}